\documentclass[final]{cvpr}

\makeatletter
\@namedef{ver@everyshi.sty}{}
\makeatother
\usepackage{tikz}
\usepackage{times}
\usepackage{epsfig}
\usepackage{graphicx}
\usepackage{amsmath}
\usepackage{amssymb}

\usepackage[normalem]{ulem}
\usepackage{siunitx}

\usepackage{bbm}  %
\usepackage{bm}  %
\usepackage{caption}
\usepackage{booktabs}
\usepackage{pifont}  %

\usepackage[pagebackref=true,breaklinks=true,letterpaper=true,colorlinks,bookmarks=false]{hyperref}

\makeatletter
\newcommand*{\transpose}{%
	{\mathpalette\@transpose{}}%
}
\newcommand*{\@transpose}[2]{%
	\raisebox{\depth}{$\m@th#1\intercal$}%
}
\makeatother

\newcommand{\affinity}[1]{\mathbf{A}_v^{#1}}
\newcommand{\attnmap}{\mathbf{A}}
\newcommand{\cnn}{f}
\newcommand{\enclayer}{g}
\newcommand{\real}{\mathbb{R}}
\newcommand{\yvosfirstvalG}{81.7}
\newcommand{\yvosfirstvalJseen}{81.2}
\newcommand{\yvosfirstvalJunseen}{76.0}
\newcommand{\yvossecondvalG}{81.8}
\newcommand{\yvossecondvalJseen}{80.9}
\newcommand{\yvossecondvalJunseen}{76.6}
\newcommand{\davisvalpretrainJnF}{82.5}
\newcommand{\davisvalpretrainJ}{79.9}
\newcommand{\davisvalpretrainF}{85.1}
\newcommand{\davisvalJnF}{78.4}
\newcommand{\davisvalJ}{75.4}
\newcommand{\davisvalF}{81.4}

\newcommand{\outputvar}{\mathbf{Y}}
\newcommand{\posenc}{\mathbf{P}}
\newcommand{\vidfeat}{\mathbf{T}}
\newcommand{\vidfeatpos}{\widetilde{\vidfeat{}}}
\newcommand{\vidinput}{\mathbf{S}}

\newcommand{\pixel}{\mathbf{p}}
\newcommand{\querytensor}{\mathbf{Q}}
\newcommand{\keytensor}{\mathbf{K}}

\newcommand{\valuetensor}{\mathbf{V}}

\newcommand{\softmax}{\mathtt{softmax}}
\newcommand{\yesmark}{\ding{51}}
\newcommand{\nomark}{\ding{55}}
\newcommand{\J}{$\mathcal{J}$}
\newcommand{\Jseen}{$\mathcal{J}_\textrm{seen}$}
\newcommand{\Junseen}{$\mathcal{J}_\textrm{unseen}$}
\newcommand{\F}{$\mathcal{F}$}
\newcommand{\Gsc}{$\mathcal{G}$}
\newcommand{\JandF}{$\mathcal{J} \& \mathcal{F}$}

\def\cvprPaperID{3473} %
\def\confYear{CVPR 2021}

\newcommand{\myparagraph}[1]{\textbf{#1 ---}}

\begin{document}

\title{SSTVOS\@: Sparse Spatiotemporal Transformers for Video Object Segmentation}

\author{Brendan Duke\textsuperscript{1,4}\thanks{Corresponding Author: \texttt{brendanw.duke@gmail.com}}
	\quad
	Abdalla Ahmed\textsuperscript{4}\quad
	Christian Wolf\textsuperscript{3}\quad
	Parham Aarabi\textsuperscript{1,4}\quad
	Graham W.~Taylor\textsuperscript{2,5}
	\and
	\textsuperscript{1}University of Toronto\and
	\textsuperscript{2}University of Guelph\and
	\textsuperscript{3}Universit\'e de Lyon, INSA-Lyon, LIRIS\and
	\textsuperscript{4}Modiface, Inc.\and
	\textsuperscript{5}Vector Institute%
}

\maketitle

\begin{abstract}
	\noindent
	In this paper we introduce a Transformer-based approach to video object
	segmentation (VOS).
	To address compounding error and scalability issues of prior
	work, we propose a scalable, end-to-end method for VOS called Sparse Spatiotemporal
	Transformers (SST).
	SST extracts per-pixel representations for each object in a video using sparse
	attention over spatiotemporal features.
	Our attention-based formulation for VOS allows a model to learn to
	attend over a history of multiple frames and provides suitable inductive bias for
	performing correspondence-like computations necessary for solving motion
	segmentation.
	We demonstrate the effectiveness of attention-based over recurrent networks in the
	spatiotemporal domain.
	Our method achieves competitive results on YouTube-VOS and DAVIS 2017 with
	improved scalability and robustness to occlusions compared with the state of
	the art.
	Code is available at~\url{https://github.com/dukebw/SSTVOS}.
\end{abstract}

\section{Introduction}

\begin{figure}
	\centering
	\includegraphics[width=1.0\linewidth]{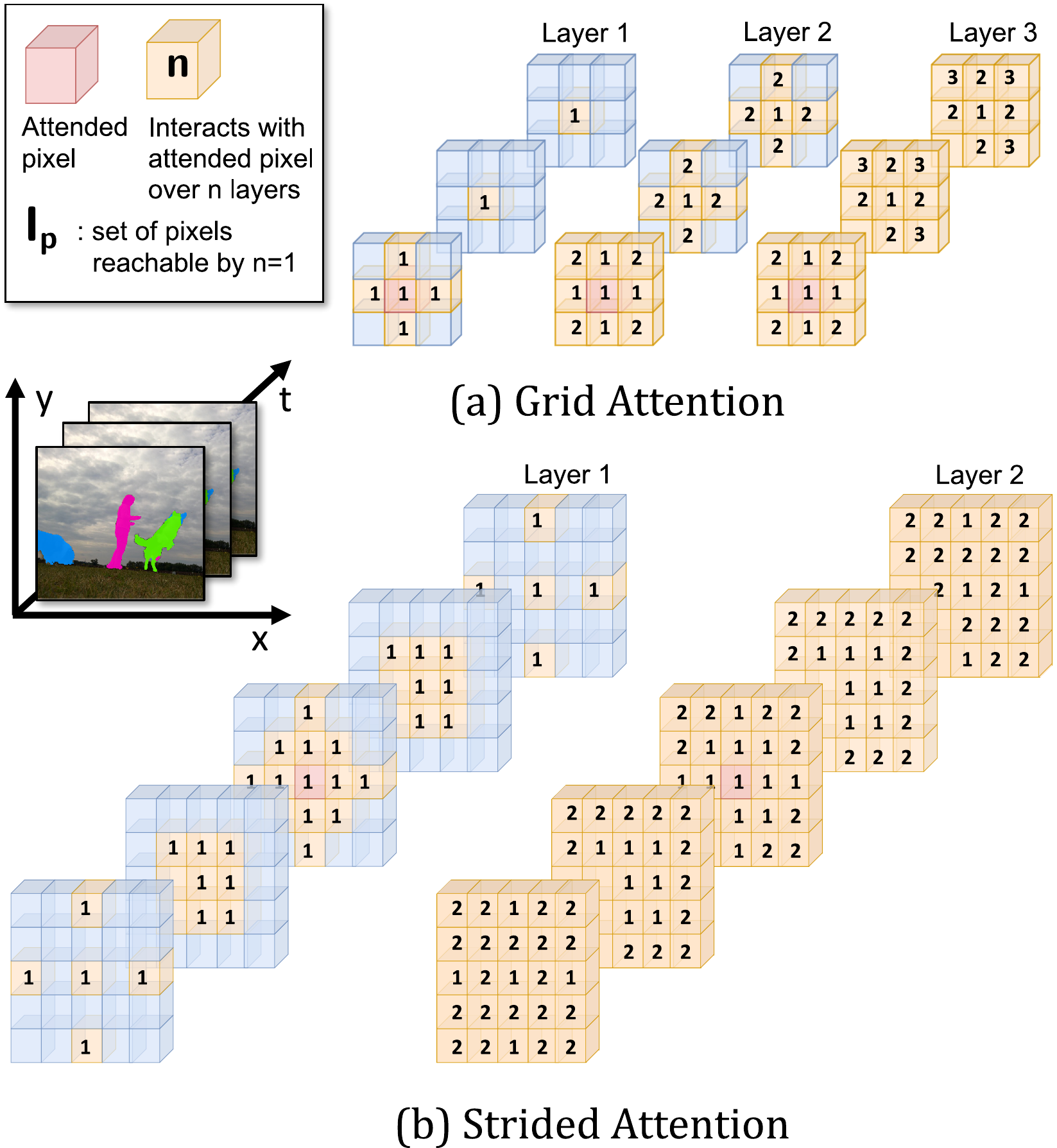}
	\caption{\label{fig:gridstridedattention}
		We propose a Transformer-based model for video object segmentation featuring
		self-attention over time and over space.
		To segment an output frame, the model learns to look up similar regions in the
		temporal history and to search for reference masks.
		We address the high computational complexity of the problem with a
		sparse Transformer formulation, which allows each cell to attend to each other
		cell over one or multiple hops.
		Here, interactions propagate from a given
		feature cell via our sparse spatiotemporal attention variants: (a) grid attention, and (b) strided attention.
	}
\end{figure}

\noindent
Video object segmentation (VOS) involves simultaneous tracking and segmentation
of one or more objects throughout a video clip.
VOS is a challenging task in which algorithms must overcome object appearance
changes, occlusion and disocclusion, as well as distinguish similar objects in
motion over time.

A highly performant VOS system is important in downstream tracking
applications where pixelwise tracking information is useful, such as player
tracking in sports analytics, person tracking in security footage, and car and
road obstacle tracking in self-driving vehicle applications.
VOS methods are also relevant in interactive annotation of video data, where
annotator time can be used more efficiently by using automatic video object
segmentation in an annotate-predict-refine loop.
Our work uses VOS as a proxy task to investigate scalable
algorithms for extracting spatiotemporal representations from video in general,
and these algorithms can be re-used for yet other video prediction tasks.

Previous methods that attempt to solve VOS can be divided into three major
categories: online finetuning, mask refinement, and temporal feature
propagation. Each of these categories, reviewed in detail in \S
\ref{sec:related}, has inherent drawbacks.
Online finetuning methods cannot adapt to changes in
object appearance throughout a sequence.
The dominant mask refinement and temporal feature propagation methods
are recurrent.
Due to their sequential nature, recurrent methods for VOS exhibit compounding
error over time, and are not parallelizable across a single example.

Motivated by the success of Transformer architectures in NLP (see \S \ref{sec:related}) we propose a novel method for semi-supervised VOS that overcomes the drawbacks
of online finetuning and sequential methods.
Our method, Sparse Spatiotemporal Transformers (SST), processes videos in a single
feedforward pass of an efficient attention-based network.
At every layer of this
net, each spatiotemporal feature vector simultaneously interacts with all
other feature vectors in the video.
SST does not require online finetuning, although it may
benefit from this practice at the cost of the aforementioned runtime penalty.
Furthermore, since SST is feedforward, it avoids the compounding error
issue inherent in recurrent methods.
Finally, SST is fully parallelizable across a single example and can therefore
take advantage of the scalability of current and future compute architectures.

Applying spatiotemporal attention operators to VOS raises two challenges:
computational complexity and distinguishing foreground objects.
Na\"ive spatiotemporal attention is square in the dimensionality of the video
feature tensor, i.e.,~$O({(THW)}^2 C)$.
We resolve this computational complexity issue with sparse attention operators,
of which we compare two promising candidates.

SST reduces feature matching FLOPs by an order of magnitude,
and achieves an overall score of~\num{\yvossecondvalG} on the official
YouTube-VOS 2019 validation set, comparing favourably with prior work.
Furthermore, we observed qualitatively improved robustness to occlusions using SST's temporal
buffer of preceding frame embeddings.

\myparagraph{Contributions}
We propose a Transformer-based model for VOS, and link
its inductive bias to correspondence calculations.
While there is work on
Transformers for representation learning in video \cite{Vilbert2019,FabienCBT2019},
these models attend over time and not densely over space.
There is also recent work that adapts Transformers to video
action recognition~\cite{girdhar2019video}, however,
we are unaware of work that uses Transformers in VOS, which requires dense predictions.
We also contribute empirical evaluation of Transformer models applied to VOS,
and argue superiority over recurrent models.

We address computational complexity using sparse attention operator variants,
making it possible to apply self-attention on high-resolution videos.
We extend sparse attention variants to video so that they can be used for VOS.
Specifically we extend to 3D, with two spatial axes and one temporal axis,
Criss-Cross Attention~\cite{huang2018ccnet} from 2D semantic segmentation, and
Sparse Attention~\cite{child2019sparsetransformer} from 1D language
translation.
Our sparse video attention operators are not VOS specific, and could be applied
to other dense video prediction tasks.
We provide our implementation~\cite{sst-github,sstvos-github}.

\section{Related Work}
\label{sec:related}

\noindent Our work is related to previous efforts in VOS. We are motivated by work on Transformer architectures in language
translation, as well as by orthogonal work on correspondence matching in
computer vision.

\myparagraph{Video Object Segmentation}
In the VOS literature, \textbf{online finetuning}
approaches~\cite{bao2018cnn,caelles2017one,maninis2018video,voigtlaender2017online,hu2018motion,khoreva2017learning,li2018video,luiten2018premvos}
do one-shot, or semi-supervised, VOS by finetuning a semantic segmentation
network on an initial frame.
A number of methods~\cite{cheng2018fast,caelles2017one,maninis2018video,voigtlaender2017online}
performed VOS on independent frames using finetuning alone without explicitly
modeling temporal coherence.
Maninis et al.~\cite{maninis2018video} built on the original usage of this
approach by Caelles et al.~\cite{caelles2017one} by adding instance-level semantic
information, while Voigtlaender and Leibe~\cite{voigtlaender2017online} added
adaptive training during the sequence.
\textbf{Offline} methods must use temporal information to produce future
segmentations from past frames, as done using optical flow
by Jang and Kim~\cite{jang2017online} and Tsai et al.~\cite{tsai2016video}.
Our method is in principle able to take advantage of online finetuning to
improve performance, and also performs competitively using offline training
alone.

Research into temporal coherence in VOS splits into two categories: approaches that
refine masks, and those that propagate temporal features.

\textbf{Mask refinement}
approaches~\cite{khoreva2017learning,oh2018fast,khoreva2017learning,hu2017maskrnn,jang2017online}
refine a previous mask using feedforward models.
Early work~\cite{khoreva2017learning} implemented mask refinement by a
recurrent connection on the concatenated frame~$t-1$ output masks and frame~$t$
RGB inputs where the recurrent connection is a VGG \cite{Simonyan2015-yf} network.
An extension concatenated the feature map from the first frame~\cite{oh2018fast}.
Yang et al.~\cite{yang2018efficient} used a spatial prior on the target
location, with a channel-wise attention mechanism and meta-learning to adapt the
network to the object given in the first frame.
Bao et al.~\cite{bao2018cnn} propagated masks by approximate inference in an
MRF, with temporal dependencies based on optical flow, and spatial dependencies
using a CNN\@.
Optical flow has also been used to add motion information via jointly training
optical flow and VOS~\cite{cheng2017segflow,jain2017fusionseg,hu2018motion}.

\textbf{Temporal feature propagation}
approaches~\cite{tokmakov2017learning,xu2018youtube,hu2017maskrnn,salvador2017recurrent,jampani2017video,johnander2019agenerative}
improve upon mask refinement by increasing the expressive power of the
mask feature representation.
At the time of writing, all such methods have used RNNs
to encode and propagate spatiotemporal representations through time.
Our approach falls under the temporal feature propagation category.
We use sparse attention operators to propagate features temporally in a single
feedforward operation.

FEELVOS~\cite{feelvos2019} is a simple and fast
method for solving the VOS problem.
Unlike most other VOS methods, FEELVOS trains end-to-end
using a pixel-wise embedding.
FEELVOS also uses global and local matching
to the reference and previous frames to predict masks for the video sequence.
Our work shares similarities with FEELVOS in
that both methods are end-to-end trainable and conceptually simple.
Our method has the added advantages of simultaneously extracting features from
multiple frames using attention, and using positional encodings to learn
spatiotemporal position-aware representations.

\myparagraph{Self-attention and Correspondence Matching}
End-to-end attention networks known as Transformers~\cite{vaswani2017attention}
are a dominant current approach to a multitude of
text natural language processing (NLP)
tasks~\cite{devlin2019bert, dai2019transformer}, vision and language \cite{tan2019lxmert}
as well as speech recognition
tasks~\cite{luscher2019rwth, kim2019improved}. Recent work has explored ties between attention heads and different reasoning functions \cite{voita2019analyzing,ReasoningCVPR2021}.
More recently, Transformers have also been applied in computer vision with success~\cite{NLNLCVPR2018,GirdharCVPR2020,ZhaoJiaKolutionCVPR2020,16X16openreview}.
In the context of VOS, we argue that self-attention also
has the potential to overcome the short-comings
of the traditionally used recurrent methods~\cite{ventura2019rvos}.
RNNs and variants are based on a Markovian assumption, where a flat
vectorial hidden state is propagated over time through the sequence.
Our Transformer based model takes a history of several frames
and reference or predicted masks as input and allows each output
region to attend to any region in the input history. This makes
the propagated representation inherently structured.

\section{Method}

\noindent
Our proposed method for VOS consists of propagating a history of $\tau$ frames over the video sequence, and allowing the model to perform spatio-temporal attention inside this history. We argue, that the proposed high-level SST architecture (\S \ref{sec:architecture}) provides inductive bias well suited for the reasoning skills required for VOS, namely computing optical flow (attending to past similar frames) and propagating reference masks over time (attending to given frames with similar appearance). We solve the challenge of computation complexity with two variants of sparse spatiotemporal attention, the ``grid'' and ``strided'' modules (\S \ref{sec:sparse-attn}).

\begin{figure*}[t]
	\centering
	\includegraphics[width=0.9\linewidth]{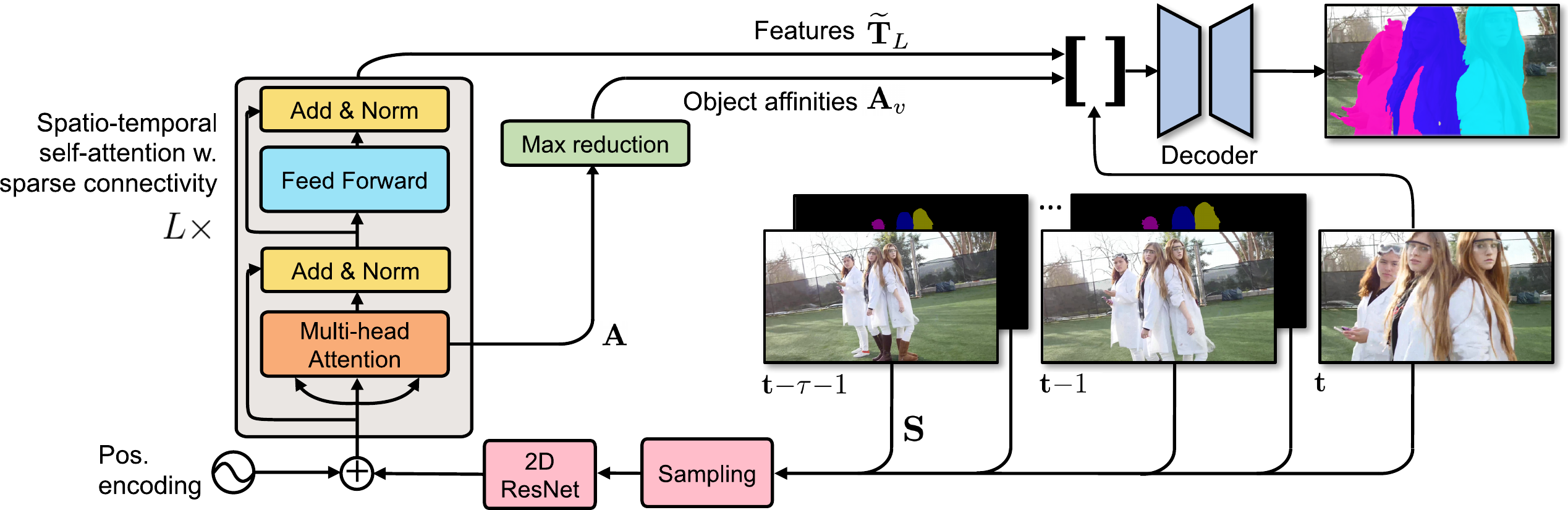}
	\caption{\label{fig:details}We propagate a history of $\tau$ frames over a video sequence and perform spatio-temporal self attention as a suitable bias for video object segmentation, allowing the model to attend to previous video frames for optical-flow like calculation, and to attend to reference frames. Computational complexity is addressed through two different sparse variants.}
\end{figure*}

\myparagraph{SST Architecture}
\label{sec:architecture}
The canonical text-based Transformer architecture~\cite{vaswani2017attention}
from which we drew motivation bears both similarities and differences with our
VOS architecture.
Like NLP Transformers, SST consists of a hierarchy of self-attention layers
that form an encoder.
In contrast to the encoder of an NLP Transformer, which takes as input embeddings
extracted from a text sequence, SST's encoder input consists of embeddings
extracted from the frames of the video to segment.
As in NLP Transformers, the SST encoder's output feeds into a decoder.
However, SST's decoder is unlike NLP Transformer decoders, which consist of
cross-attention layers that take the output sequence embeddings as input.
Instead, SST's decoder is a generic convolutional segmentation network that takes
as input a concatenated set of features: current frame embeddings, attention
values produced from SST's encoder's hierarchy of attention maps, and embedding
output by SST's encoder.
For the purpose of fair comparison with state-of-the-art work, in
\S \ref{sec:experiments} we use CFBI's decoder module~\cite{yang2020CFBI}.

The SST architecture (Fig.~\ref{fig:details}) takes a length $T$
sequence of $H \times W$ RGB video frames~$\vidinput{}\in \real{}^{3\times T\times H\times W}$
as input.
From~$\vidinput$ a CNN feature extractor~$f$ extracts
\begin{equation}
	\vidfeat{} = \cnn{}(\vidinput{}),
	\label{eqn:cnn-embedding}
\end{equation}

\noindent a per-frame embedding
$\vidfeat{}\in\real{}^{C\times T\times H'\times W'}$ at reduced spatial
resolution~$H'\times W'$ and with embedding channels~$C$.
In our experiments we used a ResNet-101 as~$f$.

In order to meet hardware resource constraints, and supposing that a given
frame's relation to past frames decreases with time, any given
frame embedding attends to a temporal buffer of its~$\tau$ preceding frame
embeddings.
Denote the truncated frame embedding buffer by~$\vidfeat{}_\tau$.
We optionally add information about the spatiotemporal
position of cells in a video tensor by the positional encoding sum
\begin{equation}
	\vidfeatpos{} = \vidfeat{}_\tau + \posenc{}
	\label{eqn:positional-encoding}
\end{equation}

\noindent where~$\posenc{} \in \real{}^{C\times T\times H'\times W'}$ encodes
absolute position.
We can encode absolute position~$\posenc{}$ using various priors, such as
sinusoids or learned embeddings~\cite{vaswani2017attention}, or as a zero
tensor in the case of no positional encoding.

The SST encoder processes positionally encoded per-frame
embeddings~$\vidfeatpos{}$ with~$L$ self-attention layers,
adding temporal context to the video representation.
The SST encoder passes two outputs to the SST decoder, the first of which is
spatiotemporal features~$\vidfeatpos{}_L\in\real{}^{C\times \tau \times H'\times W'}$.
A composition of~$L$ layers~$\enclayer_l$ computes features~$\vidfeatpos{}_L$
as
\begin{equation}
	\vidfeatpos{}_L = \enclayer{}_L\circ\enclayer_{L - 1}\circ\cdots\circ\enclayer_1(\vidfeatpos{}).
	\label{eqn:encoder-layers}
\end{equation}

\noindent Each encoder layer~$g_l$ consists of a sequence of multi-head
attention and spatiotemporal position-wise feedforward components combined with
skip connections and normalization (Fig.~\ref{fig:details}).
The output~$\vidfeatpos{}_L$ of the SST encoder feeds directly into the
decoder as the representation containing spatiotemporal context.

The SST decoder's other input arises as an intermediate tensor
computed by the multi-head attention component of each encoder layer.
Each sparse multi-head attention operation computes an attention map, which we
refer to as an object affinity tensor.
The role of the object affinity tensor is to propagate segmentation information
from past frames (either reference or predicted) to the future using the
attention distributions of the Transformer heads.
This can be seen as inductive
bias for the model allowing it to more easily tie attention to semantically
relevant motion.
Key to the procedure are tensors $I^o_\pixel$, which correspond to the pixels
in the sparse connectivity pattern of feature cell $\pixel$, and which belong to object $o$.
Connectivity pattern~$I_\pixel{}$ determines which other feature cells are
``connected to'' and thus interact with feature cell~$\pixel{}$
(Fig.~\ref{fig:gridstridedattention}).
The SST encoder uses connectivity patterns to compute the decoder's second
input: object affinity values.

Object affinity
values~$\affinity{}\in\real{}^{L\times O\times \tau\times H'\times W'}$
represent the affinity of each of the~$\tau\times H'\times W'$
cells in the spatiotemporal feature tensor with each of the~$O$ generalized
objects --- all reference objects plus the background.
Each object affinity value~$\affinity{l}(\pixel)$ is calculated as the
maximum attention score in the object affinity
tensor~$\attnmap{}^l \in \real{}^{\lvert I_\pixel{}\rvert\times \tau\times H'\times W'}$,
i.e., the score of the pixel belonging to object $o$ and most attended to
by the attention head.
Each of the~$L$ attention layers in the SST encoder computes its own object
affinity value using a reduction operation over its object affinity tensor
(Fig.~\ref{fig:details}).
To enforce causality, feature cell~$\pixel{}$'s object affinity is
computed only over previous timesteps.
Furthermore, each feature cell~$\pixel{}$ attends only to feature cells in its
connectivity pattern~$I_\pixel{}$.
We define object affinity values as
\begin{equation}
	\affinity{l}(\pixel) = \underset{I^o_\pixel{}\cup\{0\}}{\max}\attnmap{}^l
	\label{eqn:object-affinity}
\end{equation}

\noindent where~$I^o_\pixel{}$ denotes the (possibly empty) set of feature cells belonging
to object~$o$, in connectivity pattern $I_\pixel{}$, and defaulting to zero.
By taking the~$T$\textsuperscript{th} temporal slice of object affinity values~$\affinity{}$ we
obtain object-discriminative features used to infer the current frame object segmentation.

Due to the form of the multi-head attention computation described in
\S \ref{sec:sparse-attn}, attention maps~$\attnmap{}^l$ contain pairwise
dot products between feature cells and other feature cells within their
respective connectivity patterns.
From these dot products we can compute Euclidean distance or normalized cross
correlation.
Intuitively, by doing so we use the attention map features to compare the
distances or angles between per-frame embeddings in the affine subspaces
projected to by each attention head.
Taking all attention heads in the encoder together forms a hierarchy of
such distance (or angle) features.
This improves the expressiveness of the model
compared with straightforward Euclidean distance between the per-frame
embeddings~$\vidfeat{}$ from Equation~\ref{eqn:cnn-embedding}.\looseness=-1

The SST decoder (Fig.~\ref{fig:details}) is a convolutional
decoder module that takes the spatiotemporal context features~$\vidfeatpos{}_L$
and object affinity features~$\affinity{1..L}$ of all encoder layers as
input.
The final layer of the SST decoder produces the video object segmentation
probability or masks~$\outputvar{} \in \real{}^{H\times W}$ from the final
object-discriminative representation for a given frame.
It applies a scoring convolution followed by sigmoid at training time or argmax
at test time.
In the case of multiple objects we have probability scoremaps
in~$\real{}^{O\times H\times W}$, i.e.,
probabilities for each generalized object (including background) for
each pixel in the video.
An inference protocol reduces these scoremaps to a tensor
in~$\real{}^{\tau\times H\times W}$ of object integer labels.
We use the ``na\"ive'' inference protocol~\cite{oh2018fast} and take, for each
pixel, the argmax over all object probabilities including the background
probability.

\myparagraph{Sparse Attention}
\label{sec:sparse-attn}
In this section we use~$T$ to denote a generic temporal dimension, but as
described in \S \ref{sec:architecture}, we actually operate on a reduced sized
buffer of length $\tau$.

Attention is a dense operator that allows each element of a tensor to interact
with all other elements at each attention layer.
In VOS, attention can capture long-range dependencies without recurrence, and
can be viewed intuitively as a cross-correlation operator that uses CNN
features for correspondence~\cite{long2014correspondence}, similar to prior
work that used matching layers for optical flow~\cite{dosovitskiy2015flownet}.

Formally, we follow~\cite{vaswani2017attention} in defining attention as
\begin{equation}
	\mathtt{Attention}\big(\querytensor{}, \keytensor{}, \valuetensor{}\big) = \softmax{}\big(\querytensor{}\keytensor{}^{\transpose}\big)\valuetensor{},
	\label{eqn:attndefn}
\end{equation}

\noindent where query~$\querytensor{}$, key~$\keytensor{}$, and value~$\valuetensor{}$
are all matrices in~$\real{}^{S\times C}$ for flattened spatiotemporal dimensions~$S = THW$.
As we alluded to in \S \ref{sec:architecture}, we use spatiotemporal
features~$\vidfeat{}$ as query, key, and value, i.e., we do self-attention.
Intuitively we increase the spatiotemporal context of each feature
cell~$\pixel{}$ by doing a lookup in the spatiotemporal features connected
to~$\pixel{}$.

We adapted for VOS characteristic components of the Transformer architecture as
described by Vaswani et al.~\cite{vaswani2017attention}, including multi-head
attention and positional encodings.
We compare the effectiveness of positional encoding schemes applied to VOS in
\S \ref{sec:davis2017ablation}.
We did not normalize the softmax argument in Equation~\ref{eqn:attndefn} by the
inverse square root of channels, as we found this scaling factor reduced model
effectiveness.
The difference in impact of scaling factor between our VOS attention and
Vaswani et al.'s NLP attention could be due to our attention operator having a
comparatively low number of channels.

A computational barrier prevents na\"ively using Equation~\ref{eqn:attndefn} to
perform our desired self-attention operation on spatiotemporal features~$\vidfeat{}$.
The attention operation given in Equation~\ref{eqn:attndefn} is~$O({(THW)}^2
	C)$, which poses a problem for video object segmentation since for dense
prediction tasks such as segmentation, model performance tends to improve with
greater input resolution~\cite{zhao2017icnet}.
As an illustration of the infeasibility of using na\"ive attention for VOS,
consider that a single layer of attention on a~\num{16} frame video with
a~$64\times 64$ feature map with~\num{32} channels would cost more
than~\num{137} billion FLOPs, far more than the most computationally expensive
CNNs in the literature at the time of writing~\cite{tan2019efficientnet}.

We propose to use sparse spatiotemporal attention operators to overcome this
computational barrier to applying attention for VOS\@.
We define sparse attention operators using a connectivity
pattern set~$I = \{I_{\pixel{}_0}, \dots, I_{\pixel{}_S}\}$
where~$I_\pixel{}$ is a set of coordinates~$(i, j, k)$ that index a 3D tensor.
Here again, connectivity pattern~$I_\pixel{}$ determines which other feature
cells interact with feature cell~$\pixel{}$.

For query~$\querytensor{}$, key~$\keytensor{}$, and
value~$\valuetensor{}$ tensors all in~$\real{}^{C\times T\times H\times W}$,
a sparse attention operator is defined as
\begin{equation}
	\mathtt{SparseAttn}{\big(\querytensor{}, \keytensor{}, \valuetensor{})}_\pixel{} = \softmax{}\big(\querytensor{}_\pixel{} \keytensor{}^{\transpose}_{I_\pixel{}}\big)\valuetensor{}_{I_\pixel{}}.
	\label{eqn:sparseattndefn}
\end{equation}

\noindent We adapt two different sparse attention methods from 1D or 2D to 3D to make our
attention operator computationally tractable at our desired framerate and
resolution.
We achieve computational tractability by careful selection of the connectivity
patterns of our sparse attention operators.

\myparagraph{Grid Attention}
We adapted our first sparse attention operator from Huang et al., who also
noted the computational complexity issue when applying attention for semantic
segmentation~\cite{huang2018ccnet}.
In VOS, however, the computational complexity issue is exacerbated by the
addition of the time dimension.
We refer to the generalized operator described below as ``grid attention'' since the moniker
``criss-cross attention'' is no longer fitting in more than two dimensions.

At each layer of grid attention, each feature cell of the spatiotemporal feature tensor
aggregates information from other feature cells along its~$X$,~$Y$, and~$T$ axes
independently.
Each feature cell interacts once with every other feature cell in the spatiotemporal feature
tensor that shares at least two of its~$X$,~$Y$, and~$T$
coordinates.

Grid attention implements Equation~\ref{eqn:sparseattndefn} with a connectivity
pattern~$I_\pixel{}$ consisting of~$(T + H + W - 2)$ feature cell indices.
$I_\pixel{}$ contains all feature cells along the horizontal, vertical, or
temporal axes incident to location~$\pixel{} \equiv (x, y, t)$.
The grid attention
weights~$\softmax{}\big(\querytensor{}_\pixel{} \keytensor{}^{\transpose}_{I_\pixel{}}\big)$
are then a matrix in~$\real{}^{S\times (T + H + W - 2)}$, each row of which
contains weights of a convex combination.
For a feature cell~$\pixel{}$ in the spatiotemporal feature tensor, the grid
attention weights are over all feature cells along~$\pixel{}$'s temporal,
vertical, and horizontal axes.
By multiplying by the grid attention weights we attend over~$\pixel{}$'s grid
connectivity pattern.
Note that we implemented our grid attention operator in place, so we incur no
overhead from indexing tensors by~$\pixel{}$.

In Figure~\ref{fig:gridstridedattention} (top) we illustrate how grid attention propagates
interactions from a single attended feature cell to all other feature cells in three
sequential layers.
The first grid attention layer propagates information to other feature cells in the
same frame vertically and horizontally, and to feature cells at the same spatial
location in all other frames.
The second layer propagates interactions to the entire current frame, and
vertical and horizontal axes in other frames.
Finally, the third layer propagates information to all feature cells in the video
feature tensor.

We can show that composing three applications of grid self-attention on
spatiotemporal feature tensor~$\vidfeat{}$ produces an output tensor where each
spatiotemporal feature cell with coordinates~$(x, y, t)$ is composed of a weighted sum
\begin{equation}
	\sum_{i = 1}^W \sum_{j = 1}^H \sum_{k = 1}^T
	\big(\vidfeat{}^{\transpose}_{xyt} \vidfeat{}_{iyt}\big)
	\big(\vidfeat{}^{\transpose}_{iyt} \vidfeat{}_{ijt}\big)
	\big(\vidfeat{}^{\transpose}_{ijt} \vidfeat{}_{ijk}\big)\vidfeat{}_{ijk}
	\label{eqn:gridattnrouting}
\end{equation}

\noindent over other feature cells in~$\vidfeat{}$ with coordinates~$(i, j, k)$.
Equation~\ref{eqn:gridattnrouting} shows that grid attention propagates
information along ``routes'' through the spatiotemporal feature tensor.
For a feature cell at
position~$(x, y, t)$ to interact with another feature cell at an arbitrary
position~$(i, j, k)$, interactions must propagate along a ``route''  composed
of pairs of similar feature cells.
Just as we might give travel directions through a city grid such as ``first
walk ten blocks North, then walk three blocks East'', grid attention
interactions must propagate a fixed number of feature cells in the~$X$,~$Y$, and~$T$
directions, in some order, before connecting the source feature cell with
its target feature cell.

By replacing dense attention with grid attention we reduced the computational
complexity of video attention from~$O(C{(THW)}^2)$ to~$O(C(T + H + W)THW)$,
achieving our goal of making attention tractable for video.

\myparagraph{Strided Attention}
We investigated strided attention as an alternative sparse attention method in
addition to grid attention.
Drawing inspiration from sparse Transformers for sequences~\cite{child2019sparsetransformer},
information propagates by following paths of locations through sparse
connectivity patterns in the spatiotemporal feature tensor.

We define strided attention's connectivity pattern~$I_\pixel{}$ as a
generalization of Child et al.'s strided attention from 1D to 3D\@.
Our strided attention uses two different connectivity patterns~$I^1_\pixel{}$
and~$I^2_\pixel{}$ corresponding to separate, sequential heads of multihead
attention.
The first connectivity pattern~$I^1_\pixel{}$ routes to all feature cells in a cube of
side-length~$h$ from~$\pixel{}$,
i.e.,~$I^1_\pixel{} = (\pixel{} + (l_x, l_y, l_z) \,:\, l_x, l_y, l_z < h)$.
The subsequent connectivity pattern~$I^2_\pixel{}$ routes to all feature cells in the
video tensor that can reach~$\pixel{}$ by taking steps of size~$h$ along each
axis,
i.e.,~$I^2_\pixel{} = (\pixel{} + (l_x, l_y, l_z) \,:\, l_x, l_y, l_z \mod h = 0)$.
We choose~$h\approx \sqrt{H}$ to reduce the computational complexity by a
square root from~$O(C{(THW)}^2)$ to~$O(C{(THW)}^{3/2})$.
We visualize strided attention's connectivity pattern in
Figure~\ref{fig:gridstridedattention} (bottom).

\begin{table}[t] \centering
	{\small
		\begin{tabular}{@{}lccc@{}}
			\toprule
			Model             & \begin{tabular}{@{}c@{}}MACs \\ (GFLOPs)\end{tabular}
			                  & \begin{tabular}{@{}c@{}}Slowdown \\ (\%)\end{tabular}
			                  & \begin{tabular}{@{}c@{}}Params\\ (M)\end{tabular}               \\
			\midrule
			DeepLab-v3        & 255.4                          & -    & 66.5 \\
			Matching (CFBI)   & 99.6                           & 39.0 & 0    \\
			\midrule
			SST (Local)       & 5.34                           & 2.1  & 0.3  \\
			SST (Strided)     & 1.89                           & 0.7  & 0.3  \\
			SST (Grid)        & 1.45                           & 0.6  & 0.3  \\
			Na\"ive Attention & 170.1                          & 66.6 & 0.3  \\
			\bottomrule
		\end{tabular}
		\caption{\label{tab:runtime}Runtime and parameter analysis.}
	}
\end{table}

The relative efficiency of grid and strided attention depends on the size
of~$T$, since we assume that~$H$ and~$W$ are both large relative to~$T$.
In a setup where~$H, W \in \{64, 128\}$, and~$T\approx 8$, strided attention costs
about~\num{1.3} to~\num{1.4} times as many operations as grid attention.

\myparagraph{Runtime}
Table~\ref{tab:runtime} provides a runtime and parameter analysis.
We computed the multiply-accumulates (MACs) of SST for a 3-frame temporal buffer, input resolution of~$465\times 465$,
128 channels, and 3 Transformer layers.
We show MACs in both absolute GFLOPs and as slowdown relative to DeepLab-v3 backbone MACs.
We also compare to CFBI's local/global matching.
Note that SST's local temporal window ($\tau = 3$) is larger than CFBI's ($\tau = 1$).
Finally, we compare to na\"ive attention.
Both na\"ive attention and CFBI's global matching attend pairwise to an entire feature map,
costing 39.0\% and 66.6\% slowdown relative to DeepLab-v3's runtime.
In contrast, SST factorizes the computation by attending to all other spatiotemporal feature cells over sequential layers (Fig.~\ref{fig:gridstridedattention}).
SST reduces slowdown by more than an order of magnitude to~$\approx2\%$.

\begin{figure*}[t]
	\centering
	\includegraphics[width=\linewidth]{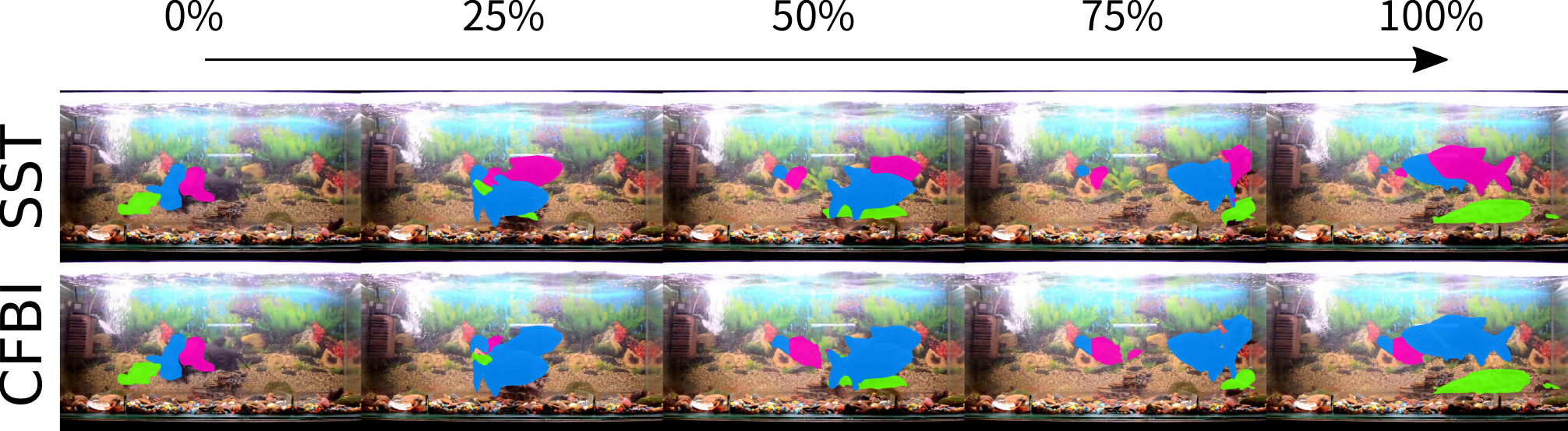}
	\caption{A qualitative example from YouTube-VOS validation showing SST handling
		occlusion.
		In this challenging example, SST's temporal history enables robust tracking of
		all three fish, while CFBI confuses two fish once they overlap.}
	\label{fig:sst-cfbi}
\end{figure*}

\section{Experiments and Results}
\label{sec:experiments}
\noindent
We present benchmark experiment results against state-of-the-art (SOTA) methods
on the DAVIS 2017~\cite{ponttuset2017davis} and
YouTube-VOS~\cite{xu2018youtubevos} datasets.
We further analyze the effect of different sparse attention operators, history sizes and
positional encodings through ablation studies on DAVIS 2017\@.

\begin{table} \centering
	{\small
		\begin{tabular}{@{}lccrrr@{}}
			\toprule
			Method                                 & O-Ft      & S         & \begin{tabular}{@{}r@{}}\Gsc{} \\ (\%)\end{tabular} & \begin{tabular}{@{}r@{}}\Jseen{} \\ (\%)\end{tabular} & \begin{tabular}{@{}r@{}}\Junseen{} \\ (\%)\end{tabular} \\
			\midrule
			\multicolumn{6}{c}{\emph{YouTube-VOS 2018 Validation Split}}                                                                                                      \\
			\midrule
			MSK~\cite{khoreva2017learning}         & \yesmark  & \nomark   & 53.1                           & 59.9                           & 45.0                           \\
			OnAVOS~\cite{voigtlaender2017online}   & \yesmark  & \nomark   & 55.2                           & 60.1                           & 46.6                           \\
			OSVOS~\cite{caelles2017one}            & \yesmark  & \nomark   & 58.8                           & 59.8                           & 54.2                           \\
			S2S~\cite{xu2018youtube}               & \yesmark  & \nomark   & 64.4                           & 71.0                           & 55.5                           \\
			PReMVOS~\cite{luiten2018premvos}       & \yesmark  & \nomark   & 66.9                           & 71.4                           & 56.5                           \\
			BoLTVOS~\cite{voigtlaender2019boltvos} & \yesmark  & \nomark   & 71.1                           & 71.6                           & 64.3                           \\
			\addlinespace[1mm]
			RGMP~\cite{oh2018fast}                 & \nomark   & \yesmark  & 53.8                           & 59.5                           & 45.2                           \\
			STM~\cite{oh2019video}                 & \nomark   & \yesmark  & 79.4                           & 79.7                           & 72.8                           \\
			KMN~\cite{seong2020kernelizedmn}       & \nomark   & \yesmark  & 81.4                           & \textbf{81.4}                  & 75.3                           \\
			\addlinespace[1mm]
			STM$^-$~\cite{oh2019video}             & \nomark{} & \nomark{} & 68.2                           & -                              & -                              \\
			OSMN~\cite{yang2018efficient}          & \nomark   & \nomark   & 51.2                           & 60.0                           & 40.6                           \\
			S2S~\cite{xu2018youtube}               & \nomark   & \nomark   & 57.6                           & 66.7                           & 48.2                           \\
			A-GAME~\cite{johnander2019agenerative} & \nomark   & \nomark   & 66.0                           & 66.9                           & 61.2                           \\
			CFBI~\cite{yang2020CFBI}               & \nomark   & \nomark   & 81.4                           & 81.1                           & 75.3                           \\
			\textbf{SST (Local)}                   & \nomark   & \nomark   & \textbf{\yvosfirstvalG}        & \yvosfirstvalJseen             & \textbf{\yvosfirstvalJunseen}  \\
			\midrule
			\multicolumn{6}{c}{\emph{YouTube-VOS 2019 Validation Split}}                                                                                                      \\
			\midrule
			CFBI~\cite{yang2020CFBI}               & \nomark   & \nomark   & 81.0                           & 80.0                           & \textbf{77.9}                  \\
			\textbf{SST (Local)}                   & \nomark   & \nomark   & \textbf{\yvossecondvalG}       & \textbf{\yvossecondvalJseen}   & \yvossecondvalJunseen          \\
			\bottomrule
		\end{tabular}
		\caption{\label{tab:sota-yvos}Comparison with SOTA methods on YouTube-VOS~\cite{xu2018youtubevos} 2018 and 2019.
			Region similarity over seen (\Jseen{}) and unseen (\Junseen{})
			categories, and overall
			score~\Gsc{} are computed as in standard benchmarks~\cite{perazzi2016abenchmark}.
			We distinguish methods by those that use online finetuning (O-Ft) and/or synthetic data (S), and
			those that do not.}
	}
\end{table}

\myparagraph{YouTube-VOS} is a large scale VOS dataset comprised
of~\num{4453} YouTube video clips spanning~\num{94} object categories~\cite{xu2018youtubevos}.
YouTube-VOS includes an official validation set of~\num{507} videos with
held out labels, which can be evaluated only through an evaluation server.
YouTube-VOS has been the basis of challenges in 2018 and 2019, yielding
two versions of the validation set and evaluation server.
The 2019 version contains new and corrected annotations.
The YouTube-VOS validation set contains~\num{26} object categories that are
unique to the validation set, used to test the generalization capability of VOS
models to object classes unseen in the training set.
The convention is to compute region similarity~\J{} and contour accuracy~\F{}
as defined by Perazzi et al.~\cite{perazzi2016abenchmark}.
As a single metric for comparing results, it is also customary to compute
overall score~\Gsc{} as the average of four values comprising region similarity
and contour accuracy scores for seen and unseen classes.

In Table~\ref{tab:sota-yvos} we present our model's results on YouTube-VOS 2018 and 2019
alongside previous SOTA results.
Our model (SST) performs favourably against all previous methods in overall
score~\Gsc{}, even methods that use online finetuning (denoted by O-Ft) and
pre-training on synthetic data (denoted by S).
Note that our unique method performs competitively against recurrent methods
that have undergone multiple research and development cycles where one method
builds on the foundation of another, for
example~\cite{johnander2019agenerative} extends~\cite{oh2018fast}, which in
turn extends~\cite{khoreva2017learning}.

\begin{table}
	\centering
	{\small
		\begin{tabular}{lccrrr}
			\toprule
			Method                                           & O-Ft     & S        & \begin{tabular}{@{}r@{}}\JandF{} \\ (\%)\end{tabular} & \begin{tabular}{@{}r@{}}\J{} \\ (\%)\end{tabular} & \begin{tabular}{@{}r@{}}\F{} \\ (\%)\end{tabular} \\
			\midrule
			\multicolumn{6}{c}{\emph{DAVIS 2017 Validation Split}}                                                                                                                    \\
			\midrule
			OSVOS-S~\cite{maninis2018video}                  & \yesmark & \yesmark & 68.0                           & 64.7                           & 71.3                           \\
			\addlinespace[1mm]
			OSVOS~\cite{caelles2017one}                      & \yesmark & \nomark  & 60.3                           & 56.6                           & 63.9                           \\
			OnAVOS~\cite{voigtlaender2017online}             & \yesmark & \nomark  & 65.4                           & 61.6                           & 69.1                           \\
			CINM~\cite{bao2018cnn}                           & \yesmark & \nomark  & 70.6                           & 67.2                           & 74.0                           \\
			PReMVOS~\cite{luiten2018premvos}                 & \yesmark & \nomark  & 77.7                           & 73.9                           & \underline{81.7}               \\
			\addlinespace[1mm]
			RGMP~\cite{oh2018fast}                           & \nomark  & \yesmark & 66.7                           & 64.8                           & 68.6                           \\
			STM$^-$~\cite{oh2019video}                       & \nomark  & \yesmark & 71.6                           & 69.2                           & 74.0                           \\
			STM$^\dagger$~\cite{oh2019video}                 & \nomark  & \yesmark & 81.8                           & 79.2                           & 84.3                           \\
			KMN~\cite{seong2020kernelizedmn}                 & \nomark  & \yesmark & 76.0                           & 74.2                           & 77.8                           \\
			\addlinespace[1mm]
			OSMN~\cite{yang2018efficient}                    & \nomark  & \nomark  & 54.8                           & 52.5                           & 57.1                           \\
			FAVOS~\cite{cheng2018fast}                       & \nomark  & \nomark  & 58.2                           & 54.6                           & 61.8                           \\
			VM~\cite{hu2018videomatch}                       & \nomark  & \nomark  & -                              & 56.5                           & -                              \\
			DyeNet~\cite{li2018video}                        & \nomark  & \nomark  & 69.1                           & 67.3                           & 71.0                           \\
			A-GAME$^\dagger$~\cite{johnander2019agenerative} & \nomark  & \nomark  & 70.0                           & 67.2                           & 72.7                           \\
			FEELVOS$^\dagger$~\cite{feelvos2019}             & \nomark  & \nomark  & 71.5                           & 69.1                           & 74.0                           \\
			CFBI~\cite{yang2020CFBI}                         & \nomark  & \nomark  & 74.9                           & 72.1                           & 77.7                           \\
			CFBI$^\dagger$~\cite{yang2020CFBI}               & \nomark  & \nomark  & 81.9                           & 79.3                           & 84.5                           \\
			\textbf{SST (Local)}                             & \nomark  & \nomark  & \underline{\davisvalJnF}       & \underline{\davisvalJ}         & \davisvalF                     \\
			\textbf{SST (Local)}$^\dagger$                   & \nomark  & \nomark  & \textbf{\davisvalpretrainJnF}  & \textbf{\davisvalpretrainJ}    & \textbf{\davisvalpretrainF}    \\
			\bottomrule
		\end{tabular}
		\caption{\label{tab:sota-davis}Comparison with SOTA methods on DAVIS 2017~\cite{ponttuset2017davis}.
			We denote online finetuning methods by O-Ft, and synthetic data methods by S\@.
			We report results trained only on the DAVIS 2017 training set, and pre-trained on YouTube-VOS.
			YouTube-VOS pre-training is denoted by~$^\dagger$.}
	}
\end{table}

\myparagraph{DAVIS 2017}
is the latest dataset in the DAVIS initiative to promote VOS
research.
DAVIS 2017 comprises 150 sequences, which include 376 separately annotated
objects~\cite{ponttuset2017davis}.
We additionally evaluate our method on DAVIS 2017~\cite{ponttuset2017davis}, and
compare our results with SOTA in Table~\ref{tab:sota-davis}.
We report our DAVIS results following the traditionally used region
similarity~\J{} and contour accuracy~\F{} metrics as well as their
mean~\JandF{}.
Our DAVIS 2017 evaluation provides additional experimental evidence that SST
performs favourably compared with existing SOTA methods, since SST achieves a
mean~\JandF{} score of~$\davisvalJnF$, whereas previous SOTA scored a~\JandF{}
of~\num{74.9} under a comparable experimental setup (without online finetuning
or synthetic data).

We evaluate only on DAVIS 2017 and not DAVIS 2016~\cite{perazzi2016abenchmark}
because DAVIS 2017 is strictly a more challenging superset of DAVIS 2016.
Furthermore DAVIS 2016 contains only single object annotations and therefore we
could make only limited evaluation of SST's ability to handle multi-object
context using DAVIS 2016.

\myparagraph{Ablation Studies}
\label{sec:davis2017ablation}
Figure~\ref{fig:sst-cfbi} shows a qualitative example on the
YouTube-VOS validation set of SST handling foreground occlusion, where
one fish entirely occludes another before the second fish becomes disoccluded
again.

We used DAVIS 2017 to perform ablation studies on interesting components of
our method, including sparse attention operators, positional encodings, and
temporal history size.
We first describe and compare different design choices for our positional
encodings.

We investigated sinusoidal positional encodings for the temporal dimension, as
used in Transformers for language translation~\cite{vaswani2017attention}.
We hypothesized that sinusoidal positional encodings would be superior to
learned positional embeddings because of the data imbalance of
temporal positions in VOS datasets, which are skewed towards lower
frame numbers.
Sinusoidal positional encodings can be interpolated or extrapolated to
generalize to underrepresented absolute frame numbers, whereas
positional embeddings have no such generalization mechanism.

\begin{table}
	{\small
		\parbox{.45\linewidth}{%
			\centering
			\begin{tabular}{lr}
				\toprule
				\begin{tabular}{@{}l@{}}Positional \\ Encoding \end{tabular} & \begin{tabular}{@{}r@{}}\JandF{} \\ (\%)\end{tabular} \\
				\midrule
				None                           & $73\pm 2$                      \\
				Sinusoidal                     & $75.6\pm 0.6$                  \\
				\bottomrule
			\end{tabular}
			\caption{\label{tab:pos-encoding}Positional encodings on DAVIS 2017 val.}
		}
		\hfill
		\parbox{.45\linewidth}{%
			\centering
			\begin{tabular}{lr}
				\toprule
				$\tau$ & \JandF{} (\%) \\
				\midrule
				1      & $75.8\pm 0.3$ \\
				2      & $76.2\pm 0.3$ \\
				3      & $76.5\pm 0.3$ \\
				\bottomrule
			\end{tabular}
			\caption{\label{tab:temporal-history}Temporal histories~$\tau$ on DAVIS 2017 val.}
		}
	}
\end{table}

We present our positional encoding results in
Table~\ref{tab:pos-encoding} using the \JandF{} score on the DAVIS 2017
validation set.
The positional encoding labeled ``None'' is our baseline attention with no
positional information, while ``Sinusoidal'' uses sinusoidal
positional encodings for all spatiotemporal dimensions~$X$, $Y$, and~$T$.
To evaluate robustness to hyperparameters, we computed the mean and variance of
\JandF{} over multiple runs for each positional encoding scheme, varying the
number of Transformer layers and the temporal history.
Sinusoidal temporal positional encodings performed best, achieving both a
higher mean score and lower variance.
The superiority of positional encodings supports our hypothesis
that information about distance-from-reference is important in positional
encodings for VOS\@.
The lower variance indicates that sinusoidal positional encodings form a robust
prior for finding correspondences in VOS\@.

In Table~\ref{tab:temporal-history} we evaluate the effect of increasing SST's
temporal history~$\tau$.
We varied the temporal history while keeping other hyperparameters fixed, and
computed the variance over multiple runs.
We observed that even a modest increase in temporal history improves the \JandF{} score.
We expect that efficiency improvements further increasing the temporal history
size will improve the effectiveness of SST's temporal history mechanism even
further.

\begin{table}
	{\small
		\centering
		\begin{tabular}{lcrrr}
			\toprule
			Sparse Attention & Layers & \JandF{} & \J   & \F   \\
			\midrule
			Grid             & 1      & 65.3     & 62.3 & 68.4 \\
			Grid             & 2      & 66.1     & 62.6 & 69.5 \\
			Grid             & 3      & 64.2     & 61.0 & 67.4 \\
			Local            & 2      & 76.2     & 72.8 & 79.5 \\
			Strided          & 2      & 69.5     & 65.7 & 73.3 \\
			LocalStrided     & 2      & 72.3     & 69.1 & 75.6 \\
			\bottomrule
		\end{tabular}
		\caption{\label{tab:sparse-attention}SST sparse attention variants on DAVIS 2017 val.}
	}
\end{table}

In Table~\ref{tab:sparse-attention} we compare the performance of SST using
different sparse attention variants.
We expected that increasing the number of layers would improve grid attention's
performance due to the increasing receptive field of each feature cell.
A larger receptive field should improve the effectiveness of the object
affinity value from multi-head attention.
Grid attention's \JandF{} score increased as expected from one to two layers,
but dropped off for three layers possibly due to overfitting.
We also expected that local attention should be effective for DAVIS 2017's fast
framerate compared to strided attention, both of which we describe in \S \ref{sec:sparse-attn}.
LocalStrided attention provides a tradeoff between the local and global context
windows of local and strided attention, respectively.
So that LocalStrided attention can attend transitively to all feature cells in
just two consecutive sparse attention layers, we set the kernel size equal to
the square root of the feature tensor width.
For fair comparison, we kept the same kernel size for all strided attention
variants.
In general, local and strided attention outperformed grid attention, showing
that local and strided connectivity patterns form superior priors for VOS\@
compared with grid attention.

\myparagraph{Discussion}
We present a method for VOS
purely based on end-to-end attention.
Future work could be analogous to Transformer
models' progression on language translation tasks, where researchers
successfully applied Transformers to increasingly long sequences.
For example, Dai et al.\ combined recurrence with attention to translate
arbitrary-length sequences~\cite{dai2019transformer}, and Kitaev et al.
introduced locality-sensitive hashing instead of dot-product attention, to
reduce computational complexity from squared to~$O(N\log N)$ while using
reversible networks to model arbitrary-length sequences with constant memory
usage~\cite{kitaev2020reformer}.
In order to evaluate VOS on long sequences the VOS community would have to
overcome a dataset creation challenge, since the current benchmark dataset
YouTube-VOS contains sequences with at most~\num{36} labeled frames, sampled
at~\num{6} fps.
We propose that future work could use interactive and semi-automatic annotation
methods, based on the existing high-quality VOS models, to create datasets with
longer and therefore more challenging sequences.

\section{Conclusions}

\noindent
We presented Sparse Spatiotemporal Transformers (SST), which, up to our knowledge,
constitutes the
first application of an entirely attention-based model for video object
segmentation (VOS).
We evaluated the positive effect of positional encodings and
the advantage of attending over a history of multiple frames, suggesting
a superiority of a spatiotemporally structured representation over the flat
hidden representation of recurrent models.
We showed that SST is capable of state-of-the-art results on the benchmark VOS
dataset YouTube-VOS, attaining an overall score of~$\mathcal{G} = \yvossecondvalG$,
while having superior runtime scalability compared with previous state of the art,
including methods based on recurrence.
We provide code~\cite{sstvos-github} to reproduce all experiments in our work, including sparse
video-attention operator implementations~\cite{sst-github}, so that the community can build on
the promising idea of using attention-based models for video.
Open challenges are the memory requirements inherent in a model with only
weak Markovian assumptions, which for the moment prevents the increase of
history size to a desirable longer extent.

\textbf{Acknowledgements ---} C. Wolf acknowledges support from ANR through grant ``\emph{Remember}'' (ANR-20-CHIA-0018) of the call \emph{``AI chairs in research and teaching''}.

\par\vfill\par

\clearpage
{\small
	\bibliographystyle{ieee_fullname}
	\bibliography{ainavos}
}

\clearpage

\section{Additional Results}

\begin{figure*}[t]
	\centering
	\includegraphics[width=\linewidth]{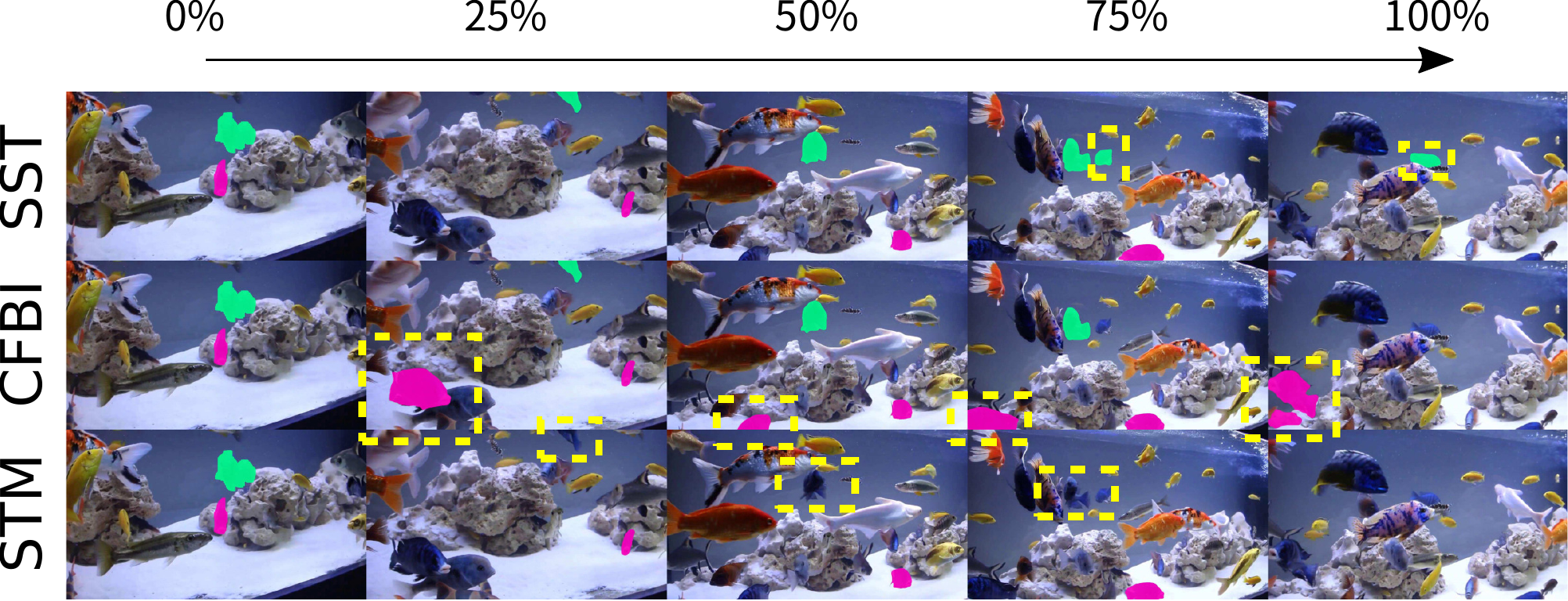}
	\caption{\label{fig:sst-cfbi-stm-fish}Fish tank. This challenging YouTube-VOS 2019 validation set example
		contains many occlusions and disocclusions by similar-looking instances of the
		same fish class.
		SST makes relatively few errors relatively later in the sequence when compared
		with CFBI~\cite{yang2020CFBI} or STM~\cite{oh2019video}.
		For clarity we labeled errors with yellow dotted boxes (best viewed digitally,
		with zoom and in colour).}
\end{figure*}

\begin{figure*}[t]
	\centering
	\includegraphics[width=\linewidth]{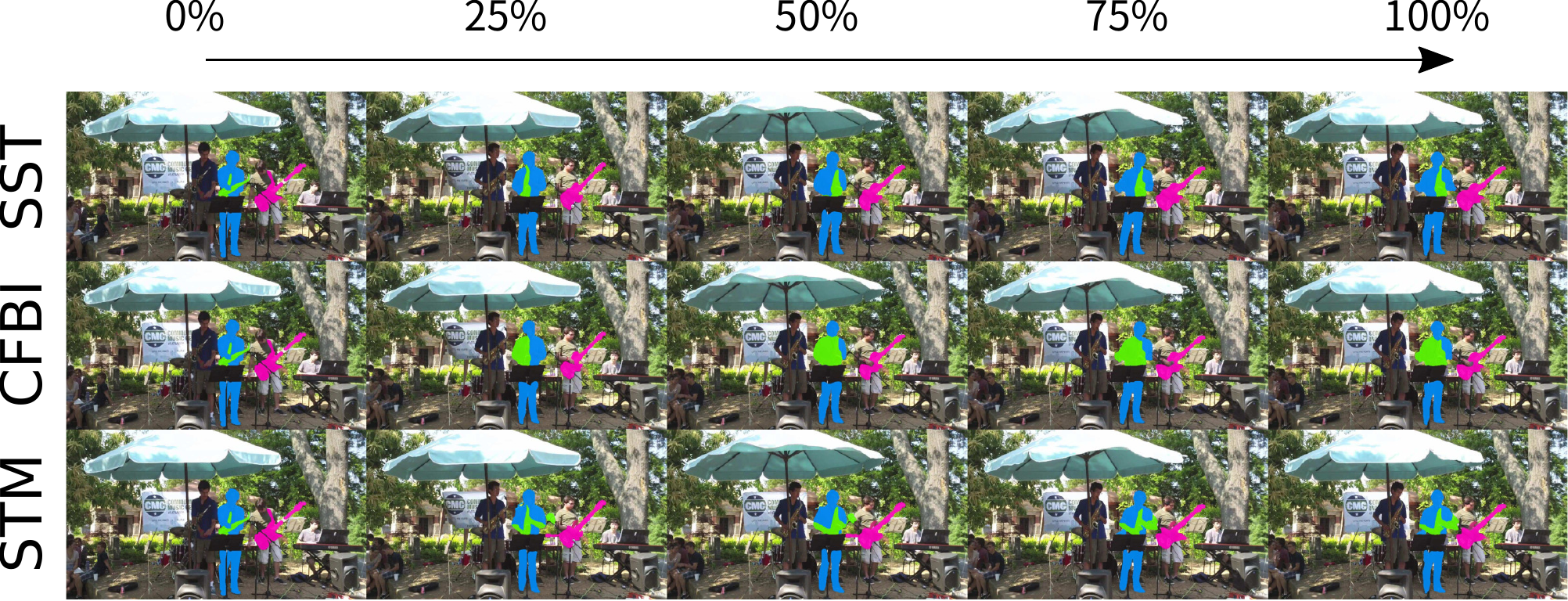}
	\caption{\label{fig:sst-cfbi-stm-jazz}Jazz band.
		In this YouTube-VOS 2019 validation set example the saxophone player
		self-occludes and disoccludes their saxophone while playing.
		SST maintains the correct saxophone segmentation throughout the sequence.
		In contrast, CFBI~\cite{yang2020CFBI} and STM~\cite{oh2019video} confuse the
		saxophone with the saxophone player's upper body after disocclusion.}
\end{figure*}

\noindent In Figures~\ref{fig:sst-cfbi-stm-fish}
and~\ref{fig:sst-cfbi-stm-jazz}, we compare SST qualitatively to
CFBI~\cite{yang2020CFBI} and STM~\cite{oh2019video}.
SST produces superior tracking in these challenging sequences, which contain
occlusions and disocclusions.
The positional encoding in the Transformer representations may enable SST to
distinguish similar instances under occlusions, using positional information.
Whereas CFBI confuses instances that are far apart, SST remains robust to these
nonlocal failures.
This further supports the effectiveness of SST's use of positional information.

\section{Grid Attention Routing}

\noindent To demonstrate mathematically information propagation in grid
attention we consider, for sake of clarity, a sparse attention function
\begin{equation}
	\mathtt{SparseAttn}{\big(\querytensor{}, \keytensor{}, \valuetensor{})}_\pixel{} = \querytensor{}_\pixel{} \keytensor{}^{\transpose}_{I_\pixel{}}\valuetensor{}_{I_\pixel{}}.
\end{equation}

\noindent where we replace the softmax by an identity function.
Further assume that our query, key, and value all are our video feature tensor,
i.e.,~$\querytensor{} = \vidfeat{}$,~$\keytensor{} = \vidfeat{}$,
and~$\valuetensor{} = \vidfeat{}$.
The first layer outputs, for each pixel~$\pixel{}$,
\begin{equation}
	\begin{split}
		\mathtt{GridAttn}{\big(\querytensor{}, \keytensor{}, \valuetensor{})}_\pixel{}_{xyt} =
		& \sum_{i = 1}^W \big(\vidfeat{}^{\transpose}_{xyt}\vidfeat{}_{iyt}\big)\vidfeat{}_{iyt} + \\
		& \sum_{\substack{j = 1 \\ j \neq y}}^H \big(\vidfeat{}^{\transpose}_{xyt} \vidfeat{}_{xjt}\big)\vidfeat{}_{xjt} + \\
		& \sum_{\substack{k = 1 \\ k \neq t}}^T \big(\vidfeat{}^{\transpose}_{xyt} \vidfeat{}_{xyk}\big)\vidfeat{}_{xyk}.
	\end{split}
\end{equation}

We can show that composing three applications of self-attention
on~$\vidfeat{}$, which we denote for brevity as~$\mathtt{GridAttn}^3$, produces
\begin{equation}
	\begin{split}
		{\mathtt{GridAttn}^3\big(\vidfeat{}, \vidfeat{}, \vidfeat{}\big)}_{xyt} =
		\sum_{i = 1}^W \sum_{j = 1}^H \sum_{k = 1}^T
		& \big(\vidfeat{}^{\transpose}_{xyt} \vidfeat{}_{iyt}\big) \\
		& \big(\vidfeat{}^{\transpose}_{iyt} \vidfeat{}_{ijt}\big) \\
		& \big(\vidfeat{}^{\transpose}_{ijt} \vidfeat{}_{ijk}\big)\vidfeat{}_{ijk} + \cdots,
	\end{split}
	\label{eqn:gridattnrouting}
\end{equation}

\noindent where~$\cdots$ represents other similar third order terms.
We show in Equation~\ref{eqn:gridattnrouting} that grid attention propagates
information along ``routes'' through the video feature tensor: for a pixel at
position~$(x, y, t)$ to interact with another pixel at an arbitrary
position~$(i, j, k)$, interactions must propagate along a ``route''  through
the video feature tensor of pairs of similar pixels.
Just as we might give travel directions through a city grid such as ``first
walk ten blocks North, then walk three blocks East'', grid attention
interactions must propagate a fixed number of pixels in the~$X$,~$Y$ and~$T$
directions, in some order, before connecting the interaction source pixel with
its target pixel.

Consider what happens if we replace the value~$\vidfeat{}_{ijk}$ returned by
the inner cross-attention in Equation~\ref{eqn:gridattnrouting} with a
foreground mask value~$m_{ijk}$.
We see that the output routes reference mask values~$m_{ijk}$ over paths of
feature vectors in the video tensor~$\vidfeat{}$ that transitively correspond
to reference features~$\vidfeat{}_{ijk}$.

\end{document}